\documentclass[nohyperref]{article}

\usepackage{microtype}
\usepackage{graphicx}
\usepackage{subfigure}
\usepackage{booktabs} 

\usepackage{hyperref}

\usepackage[accepted]{icml2022}

\usepackage{amsmath}
\usepackage{amssymb}
\usepackage{mathtools}
\usepackage{amsthm}
\usepackage{bbm}
\usepackage[capitalize,noabbrev]{cleveref}

\usepackage{enumitem}

\theoremstyle{plain}
\newtheorem{theorem}{Theorem}

\theoremstyle{definition}

\theoremstyle{remark}
\newtheorem{remark}[theorem]{Remark}

\usepackage[textsize=tiny]{todonotes}

\icmltitlerunning{Certified Robustness Against Natural Language Attacks by Causal Intervention}

\begin{document}

\twocolumn[
\icmltitle{Certified Robustness Against Natural Language Attacks\\ by Causal Intervention}




\icmlsetsymbol{equal}{*}

\begin{icmlauthorlist}
\icmlauthor{Haiteng Zhao}{pku,equal}
\icmlauthor{Chang Ma}{pku,equal}
\icmlauthor{Xinshuai Dong}{cmu,equal}
\icmlauthor{Anh Tuan Luu}{ntu,cor}
\icmlauthor{Zhi-Hong Deng}{pku}
\icmlauthor{Hanwang Zhang}{ntu}
\end{icmlauthorlist}

\icmlaffiliation{pku}{Peking University }
\icmlaffiliation{cmu}{Carnegie Mellon University}
\icmlaffiliation{ntu}{Nanyang Technological University}
\icmlaffiliation{cor}{Corresponding Author}

\icmlcorrespondingauthor{Anh Tuan Luu}{anhtuan.luu@ntu.edu.sg}

\icmlkeywords{Machine Learning, ICML}

\vskip 0.3in
]



\printAffiliationsAndNotice{\icmlEqualContribution} 

\begin{abstract}
Deep learning models  have achieved great success in many fields,
 yet they are vulnerable to adversarial examples.
This paper follows a causal perspective to look into the adversarial  vulnerability
and proposes Causal Intervention by Semantic Smoothing (CISS), a novel framework towards robustness against natural language attacks.
Instead of merely fitting observational data, CISS learns causal effects $p(y|do(x))$
 by  smoothing in the latent semantic space
 to make robust predictions, which scales to deep architectures and avoids tedious construction of noise customized for specific attacks.
CISS is provably robust against word substitution attacks, as well as empirically robust even when perturbations are strengthened by unknown attack algorithms.
%
For example, 
on YELP, CISS surpasses the runner-up by  6.8\% in terms of certified robustness against  word substitutions,
and achieves 80.7\% empirical robustness when syntactic attacks are integrated.
\end{abstract}

\section{Introduction}
\label{intro}

Deep learning models have  achieved great success in many fields  such as computer vision, natural language processing,
and speech recognition 
\citep{goodfellow2016deep,krizhevsky2012imagenet,ren2015faster,sutskever2014sequence,hinton2012deep}.
However, they are known to be vulnerable
to adversarial examples 
\citep{szegedy2013intriguing,goodfellow2014explaining,jia2017adversarial},
\emph{e.g.},
a BERT-based  sentiment analysis model can be easily fooled by synonym substitution attacks \citep{alzantot2018generating},
and thus raise severe security challenges to modern NLP  systems.
%

\begin{figure}[t]
    \centering 
    \vspace{0mm} 
    \includegraphics[width=0.85\linewidth]{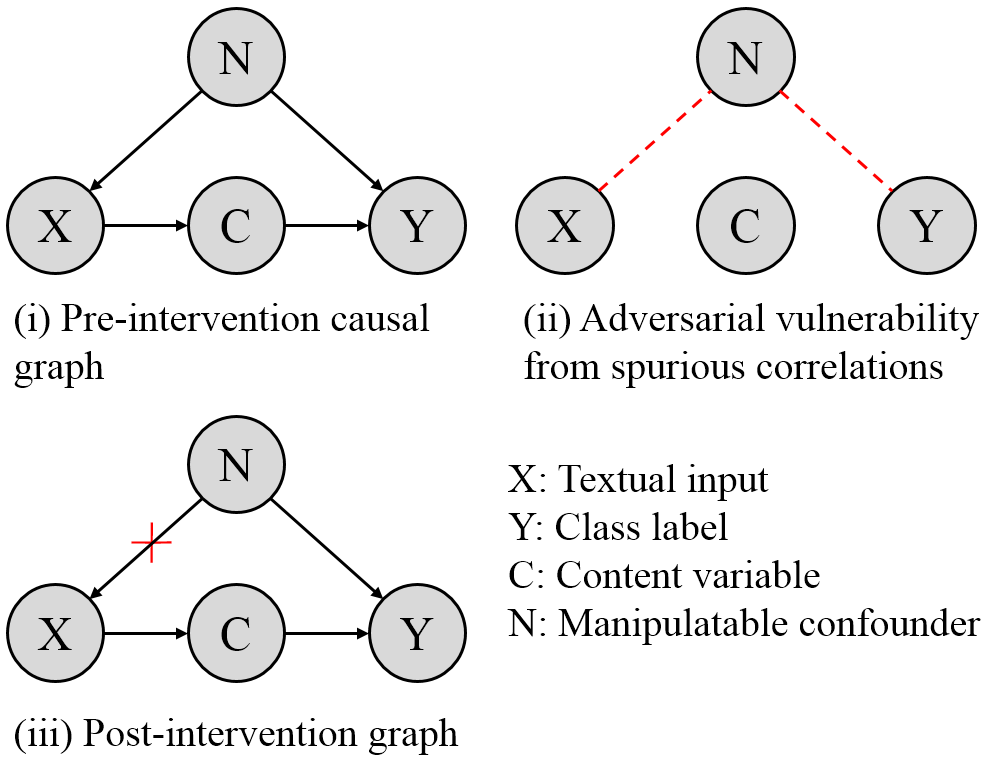}
    \vspace{-3mm} 
    \caption{
    %
    %
    %
    $N$ is a set of variables that
    can be manipulated under attacks, e.g., style variables.
    (i): The proposed causal graph of data.
    (ii): The red dashed lines represent the spurious correlation between  $X$ and $Y$,
    and it is the source of vulnerability (if a model only fits the observational distribution $p(y|x)$).
    %
    %
    (iii): The causal graph after removing the arrow from $N$ to $X$. 
    The post-intervention conditional distribution $p^{N\nrightarrow X}(y|x)$ is robust as it does not rely on any spurious correlation.
    }
    %
    \vspace{-2mm}
    \label{causal_graph}
\end{figure}

In contrast to deep models,  
humans' cognitive systems are robust against adversarial perturbations \citep{szegedy2013intriguing,goodfellow2014explaining},
as humans can perform causal reasoning \citep{pearl2009causal,peters2017elements} and 
are more sensitive to causal relations than statistical correlations \citep{gopnik2004theory}.
To enable machine learning models with such abilities to predict robustly,
it is therefore crucial to understand the adversarial vulnerability from a causal perspective \citep{zhang2021adversarial,tang2021adversarial}.

In this paper,
we consider the source of adversarial vulnerability as the 
spurious correlations by the confounder effect.
We give an illustration in Fig. \ref{causal_graph},
where 
  $X$  denotes the textual input,
  $Y$ is the class label, 
  $C$ represents the content of $X$,
  and $N$ is a set of confounder variables that
    can be manipulated under attacks.
    For example, on IMDB dataset \citep{maas2011learning},
    a professional reviewer may tend to use many jargons ($N$) in a movie review ($X$), while likely to be strict and give negative
    comments (Y).
    %
    Such correlations are useful under the i.i.d. setting but harmful if the confounder is manipulatable under attacks;
    \emph{e.g.}, adding more jargons to a positive movie review might fool a machine model.

From that causal perspective,
we draw a link between causal intervention and randomizd smoothing \citep{lecuyer2019certified,cohen2019certified}.
Randomized smoothing is a promising technique towards certified  robustness.
It adds
random Gaussian noise to the input of a base classifier and 
predicts by taking the expectation over the noise; the resulting smoothed classifier 
is robust against $l_2$-bounded perturbations with certification.
We found that  a randomized classifier actually models the causal effect $p(y|do(x))=\int p(y|x,n) p(n) \, dn$ \citep{pearl2009causal},
where $p(y|x,n)$ is the base classifier and $p(n)$ follows a Gaussian distribution.
This justifies randomized smoothing from a causal point of view
and informs us how to design and improve randomized smoothing towards robustness in certain scenarios.

Following this line of thought, we propose a novel framework towards robustness against natural language attacks, Causal Intervention by Semantic
Smoothing (CISS) (illustrated by Figure \ref{overview}).
CISS models interventional distribution $p(y|do(x))$ to predict  robustly,
by removing  the confounder effect in the semantic space.
CISS has the following merits:
(i) It has clear causal interpretation and causality-guided learning objectives;
(ii) It
provides  certified NLP robustness that  scales to deep architectures like Transformers \citep{vaswani2017attention},
while other robust certification methods like interval bound propagation \citep{jia2019certified} cannot;
(iii) It smoothes in the latent semantic space, which frees us from tedious construction of noise distributions customized for specific attacks;
(iv) In addition to certified robustness against seen attacks, CISS is empirically robust even when the perturbations are strengthened by unseen attacks.

We validate our merits by  extensive experiments considering both seen word substitution attacks \citep{jia2019certified,dong2021towards}
 and unseen syntactic-trigger-based \citep{qi2021hidden} and editing distance-based \citep{levenshtein1966binary,liang2017deep} attacks. 
For example, on IMDB, CISS achieves 76.5\% certified robust accuracy against  adversarial word substitutions,  surpassing the runner-up by 7.2\%;
on YELP, 
CISS achieves 83.1\% empirical robustness against integrated attacks, surpassing the runner-up by 7.8\%.

We  summarize the  contributions of this paper as follows:\vspace{-3mm}

\begin{itemize}
\setlength\itemsep{-0.1em}
 \item 
 We propose a causal view to look into  robustness:
   adversarial vulnerability comes from the confounding effect manipulatable by attacks,
  and 
 randomized classifiers model the causal effect only and  thus are robust to such manipulations.
 \item We propose  a novel framework, CISS, to achieve robustness against natural language attacks.
  It learns causal effects $p(y|do(x))$ to predict robustly  by smoothing in the latent semantic space.
 \item We validate that CISS is certifiably robust against known attacks and empirically robust against integrated attacks by  experiments,
 where CISS consistently surpasses the runner-up with significant margins.
\end{itemize}

    \section{Preliminaries}

    \subsection{Notations and Problem Setting}

    We suppose random variables $X, Y\sim p_{XY}(x,y)$,
    where $X\in \mathcal{X} $ represents the textual input, $Y \in  \mathcal{Y}$ represents the class label, 
     $p_{XY}$ is the  data distribution (we will use $p$ in the rest of this paper for notation simplicity),
     and $x,y$ are the observed values.
    We are interested in  a classifier $q(y|x)$ that is robust against adversarial examples.
    Given a data point $x$, an adversarial example of it,  $\hat{x} \in \mathbb{B}_{\text{adv}}(x)$,
    aims to fool a classifier, while 
    %
    $\mathbb{B}_{\text{adv}}(x)$ is defined as a neighbourhood near  $x$ to make sure that that $\hat{x}$ shares the same label with $x$ from a human's perspective.

    This paper focuses
    on  robustness against natural language  attacks, 
    which can be categorized into char-level modifications \citep{belinkov2017synthetic,gao2018black,eger2019text},
    word-level substitutions \citep{alzantot2018generating,ren2019generating,dong2021towards}, and sentence-level manipulations \citep{liang2017deep,jia2017adversarial,iyyer2018adversarial}. 
    %
    \emph{E.g.}, under adversarial word substitutions, $\mathbb{B}_{\text{adv}}(x)$ is defined as 
    $\mathbb{B}_{\text{adv}}(x) = \{\hat{x}: \hat{x}^i \in \mathbb{S}_{\text{adv}}(x^i)\}$,
    where $x^i$ is the $i^{th}$ word of $x$ and  $\mathbb{S}_{\text{adv}}(x^i)$ is a 
    pre-defined set consisting of semantically similar words of $x^i$.
    %
    %
    A classifier $q$ is said to be empirically robust at a $(x, y)$, if it 
    predicts correctly given some $\hat{x}$ that are maliciously generated by some attack algorithms,
    while a classifier $q$ is certifiably robust, if it can be theoretically guaranteed that $q$ predicts correctly given any $\hat{x}$,
    \emph{i.e.},
    $\arg \max_{y'} q(y'|\hat{x}) = y,~\forall~\hat{x} \in \mathbb{B}_{\text{adv}}(x)$.

    \section{Methodology}
    \subsection{Robustness from A Causal View}
    \label{causal perspective}

    In causal inference \citep{pearl2009causal,peters2017elements},
    observed data follow a generation process.
    This process is depicted by a set of structural equation models (SEMs)
    \citep{aldrich1989autonomy,hoover2008causality,pearl2009causal,peters2017elements},
    with a corresponding causal graph.
    To understand the source of adversarial vulnerability, 
    we  propose a  causal graph for text classification tasks, shown in 
    Fig. \ref{causal_graph} (i).
    As demonstrated,
    $X$ and $Y$ denote the textual input and the class label respectively,
    $C$ represents the content of $X$,
    and
     $N$ is a set of confounder variables that do not directly change the semantic content of the input, \emph{e.g.}, styles.
     %
    %

    Given the proposed causal graph, we are able to identify the vulnerability of a machine learning model:
    there exist spurious correlations $X\leftarrow N \rightarrow Y$, which are established by 
    confounders $N$, and such non-causal correlations can be exploited by attacks to fool a model.
    First, these spurious correlations are useful under the i.i.d. setting.
    As shown by the red dashed lines in Fig. \ref{causal_graph} (ii),
    %
     the path $X\leftarrow N \rightarrow Y$ can be used to predict $Y$.
    \emph{E.g.}, professional reviewers tend to use more jargons (N) in their review (X) and they tend to give strict negative comments (Y), and thus chances are high that a review is negative if there exist many jargons.
    Second, 
    a model which only fits the observational distribution $p(y|x)$ 
    tends to learn such features for predictions,
 not only     because such non-causal features are useful to minimize the training loss, but also because some of these non-causal features are easier to extract compared to causal relations \citep{ilyas2019adversarial,geirhos2020shortcut,tang2021adversarial}.
    Unfortunately, 
    when the confounder $N$ is manipulatable by attack algorithms,
    such spurious correlations 
    can become useless or even harmful. 
    %
    %
    \emph{E.g.}, if we employ 
 word substitution attacks to substitute the words of
 a positive movie review  by more jargons,
 a model which relies on jargons to predict is likely to give a wrong prediction.

    To learn a robust model, we therefore need to learn the causal effect from $X$ to $Y$ for prediction, 
    instead of  fitting the observational distribution $p(y|x)$.
    This can be achieved by causal interventions \citep{pearl2009causal,peters2017elements}.
    Consider a world that follows the same SEMs as Fig. \ref{causal_graph} (i), but the arrow from $N$ to $X$ is removed by intervention;
    the data from this post-intervention world follows $p^{N\nrightarrow X}(x,y,n,c)$,
    whose causal graph is shown in Fig. \ref{causal_graph} (iii).
    According to Backdoor Adjustment \citep{pearl2009causal},
    we have the following theorem (the proof of which is in the Appendix), 
    by which  the causal effect $p(y|do(x))$ equals $p^{N\nrightarrow X}(y|x)$,
    and $p^{N\nrightarrow X}(y|x)$ can be calculated by observational information from $p(x,y,n,c)$.

    %
    %
    %
    %
    %
    %

    \begin{theorem} 
    \label{theorem:causal}
        (Backdoor Adjustment \citep{pearl2009causal})
        Given $p(x,y,n,c)$  and $p^{N\nrightarrow X}(x,y,n,c)$,
        we have:
        \begin{align}
        p(y|do(x)) = p^{N\nrightarrow X}(y|x) = \int p(y|x,n)  p(n) \, dn.
        \end{align}
        \end{theorem}


    This theorem informs us that to learn a robust classifier  against manipulations on $N$, 
    we need to model $p(y|do(x))$ by modeling $p(y|x,n)$ and $p(n)$.
     We next show  this objective aligns with randomized smoothing techniques \citep{lecuyer2019certified,cohen2019certified} towards robustness.


    \begin{figure}[t]
    \centering
        \vspace{-0pt} 
        \includegraphics[width=0.95\linewidth]{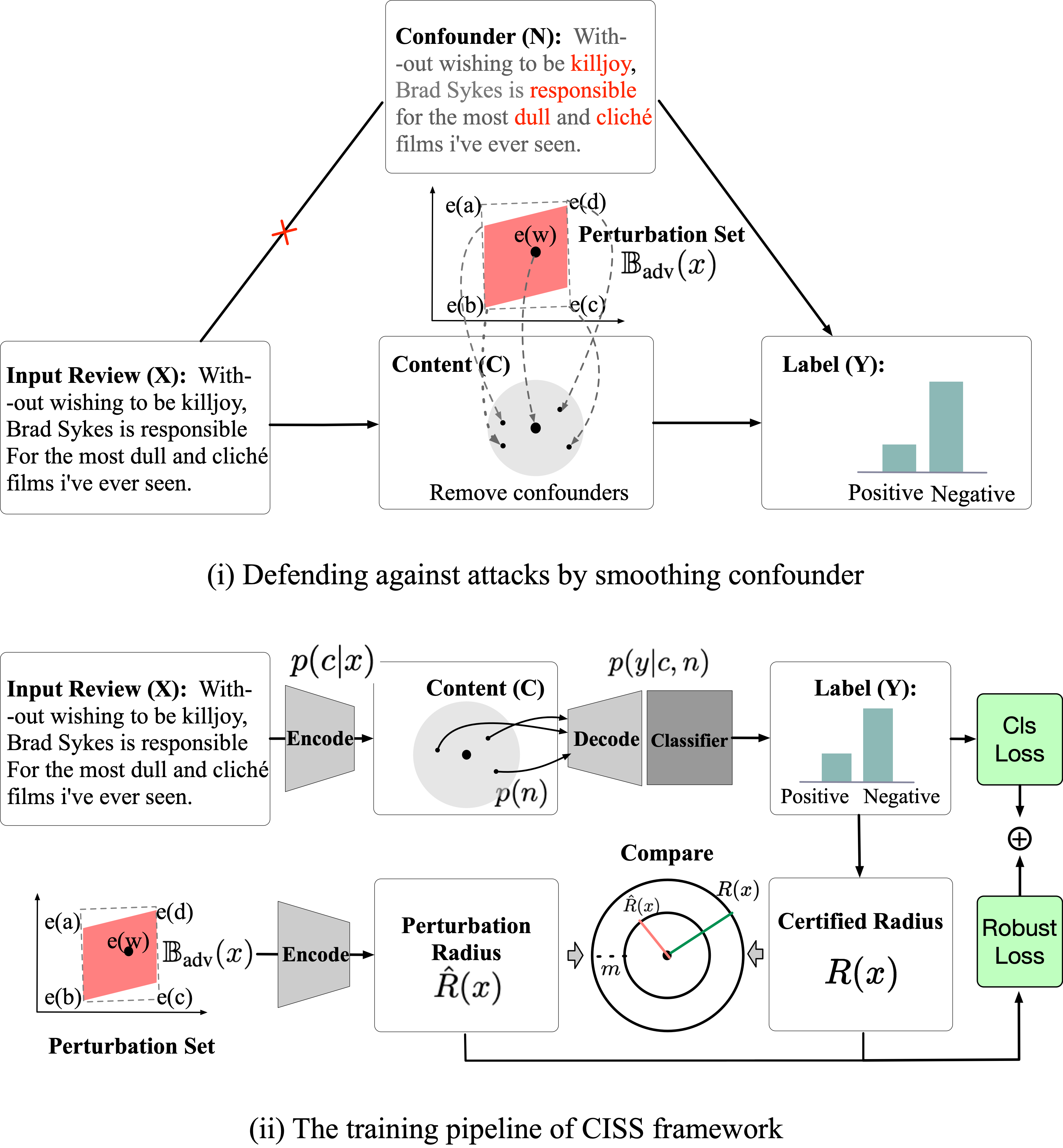}
        \caption{Overview of the proposed framework. 
        (i) Noise added in the latent space can remove the confounder effects manipulated by potential attackers.
        (ii) An illustration of the  training process of CISS.}
        \label{overview}
    \end{figure}
    
    \subsection{Randomized Classifier Captures the Causal Effect}
    \label{sec:randomized captures CE}

    Let us implement $p(y|x,n)$ by a base classifier $f_y(x+n)$, 
    and assume $p(n)$ follows $N(0,\sigma^2I)$.
    For $l_2$-bounded adversarial perturbations, 
    we have the following theorem:
    
    \begin{theorem}(Soft Smoothed Classifier \citep{Zhai2020MACERAA})
        \label{softsmooth}
        Given a base classifier  $f(x+n)$ and $p(n) \sim N(0,\sigma^2I)$.
        If 
        \begin{align}
            \mathop{\mathbb{E}}_{n\sim p(n)}[f_y(x+n)]\geq \max_{y'\neq y} \mathop{\mathbb{E}}_{n\sim p(n)}[f_{y'}(x+n)],
        \end{align}
        then 
        classification by
        $\arg\max_{y'} \mathop{\mathbb{E}}_{n\sim p(n)}[f_{y'}(\hat{x}+n)]$
        is robust for all $\hat{x}$, s.t., $\|x-\hat{x}\|_2\leq R$,
         where $R=\frac{\sigma}{2}(\Phi^{-1}(\mathop{\mathbb{E}}_{n\sim p(n)}[f_y(x+n)])-\Phi^{-1}(\max_{y'\neq y} \mathop{\mathbb{E}}_{n\sim p(n)}[f_{y'}(x+n)])$,
         and $\Phi^{-1}$ is the inverse of standard Gaussian c.d.f.
        \end{theorem}
    
        As  in  Theorem \ref{softsmooth},
         a randomized classifier adds Gaussian noise $n$ to smooth the input of a base classifier $f$ and 
         predicts the final  label by taking the expectation  over $n$.
         If $f_y(x+n)$ mimics $p(y|x,n)$ well,
         then the resulting smoothed classifier $\mathop{\mathbb{E}}_{n\sim p(n)}[f_{y}(x+n)]$
        is actually modeling the causal effect $= p(y|do(x)) = \int p(y|x,n)  p(n) \, dn$ defined in Theorem \ref{theorem:causal}.
        %

        \subsection{Modeling Causal Effect by Semantic Smoothing}
        \label{sec:semantic smoothing}

        As shown in Section \ref{sec:randomized captures CE}, the randomized smoothing technique can be used to model the interventional distribution $p(y|do(x))$; \emph{i.e.}, by assuming $p(n)$ as gaussian, the smoothed classifier $\mathop{\mathbb{E}}_{n\sim p(n)}[f_{y}(x+n)]$ captures the causal effect and is robust against $l_2$-bounded attacks.
        However, it cannot directly transfer to a NLP scenario,
        as NLP perturbations are neither continuous nor $l_2$ bounded.
        %
        As such,
        we propose a novel framework, CISS, to model the causal effect by smoothing in the latent semantic space. Therefore the random noise  can be modeled in a more flexible fashion.

        Specifically, CISS first maps the input $x$ into a semantic space by an encoder $s(\cdot)$ and then adds random gaussian noise to $s(x)$. The prediction is made by taking the expectation over $n$ and therefore the final smoothed classifier is formulated as: $\mathop{\mathbb{E}}_{n\sim p(n)}[f_{y}(s(x)+n)]$.
        %
        %
        The smoothed classifier built by CISS  has provable robustness against attacks with theoretical guarantees, which is summarized in the  following theorem (the proof of which is in the Appendix).

    
         \begin{theorem}(Robustness by Semantic Smoothing)
            \label{semanticsmooth}
            Given a base classifier $f_y(s(x)+n)$, where $s(\cdot)$ denotes an encoder that maps input into a semantic space
            and $p(n) \sim N(0,\sigma^2I)$.
            If 
            \begin{align}
                \mathop{\mathbb{E}}_{n\sim p(n)}[f_y(s(x)+n)]\geq \max_{y'\neq y} \mathop{\mathbb{E}}_{n\sim p(n)}[f_{y'}(s(x)+n)],
            \end{align}
            and  $s(\cdot)$ satisfies
            \begin{align}
                \|s(x)-s(\hat{x})\|_2\leq \hat{R}, \forall \hat{x} \in \mathbb{B}_{\text{adv}}(x),
            \end{align}
            then 
            classification by
            $\arg\max_{y'} \mathop{\mathbb{E}}_{n\sim p(n)}[f_{y'}(s(\hat{x})+n)]$
            is robust for all $\hat{x}$ if  $\hat{R}\leq R$,
            where
            \begin{equation}
            \begin{aligned}
                R=&\frac{\sigma}{2}(\Phi^{-1}(\mathop{\mathbb{E}}_{n\sim p(n)}[f_{y}(s(x)+n)])\\&-\Phi^{-1}(\max_{y'\neq y} \mathop{\mathbb{E}}_{n\sim p(n)}[f_{y'}(s(x)+n)])
                \end{aligned}
            \end{equation}
             and $\Phi^{-1}$ is the inverse of standard Gaussian c.d.f.
             
            \label{latent certi soft}
            \end{theorem}

        Similar to Theorem \ref{softsmooth}, CISS also has clear interpretability from a causal perspective.
        The  smoothed classifier $\mathop{\mathbb{E}}_{n\sim p(n)}[f_{y}(s(x)+n)]$ also
        models the  interventional distribution $p(y|do(x))$ that captures the causal effect from $X$ to $Y$, as: $
                p(y|do(x)) =     p^{N\nrightarrow X}(y|x) 
                = \int \int  p(c|x) p(y|c,n)  p(n) \, dcdn,
            $
            where $p(c|x)$ puts a point mass on $s(x)$,
            and $p(y|c,n)$ is implemented as $f_{y}(s(x)+n)$.

        Moreover, CISS has the following three merits compared to previous randomized smoothing techniques:
        (1) 
        we are able to use the flexible and tractable Gaussian noise to smooth out discrete natural language perturbations;
        (2) we are able to find the optimal trade-off between robustness and accuracy, without tuning  the hyper-parameter $\sigma$;
        (3) in addition to certified robustness, we can provide some empirical robustness even when the perturbations are strengthened by unknown  attacks algorithms.
        These merits will be detailed in the following sections and empirically validated by our experiments.

    \subsection{The Training Objective of CISS}
    In this section we introduce the training objective of CISS in detail.
    %
    %
    %
    %
     Theorem \ref{semanticsmooth} informs us how to certify the robustness at a data point $(x,y)$ by randomized smoothing in 
    the latent semantic space,
    but Theorem \ref{semanticsmooth} does not directly guide us about how the training objective should be formulated.
    To derive a training objective for CISS,
    we need to look into the interventional distribution $p(y|do(x)) = \int \int  p(c|x) p(y|c,n)  p(n) \, dcdn$,
    and formulate the training objective of CISS by the following three parts.

    \textbf{1. Training The Base Classifier.} The conditional distribution $p(y|c,n)$ corresponds to the base classifier $f_y(s(x)+n)$ in Theorem \ref{semanticsmooth}. 
    Therefore, we train $f_y(s(x)+n)$ by the classical cross-entropy loss, and take the expectation over the observational data distribution $p(x,y)$, as:
    \begin{align}
        \label{eq:lcls}
        \mathcal{L}_{\text{cls}}=  \mathop{\mathbb{E}}_{x,y\sim p(x,y)} \mathop{\mathbb{E}}_{n\sim p(n)} -\log  f_y(s(x)+n).
    \end{align}

    \textbf{2. Semantic Smoothing.} 
    %
    As for $p(c|x)$, because $p(c|x)$ 
    is defined according to the functional relationship between $c$ and $x$ by  putting a point mass on $s(x)$,  there is no need to derive an explicit training objective for it.

    However,
    as we assume $p(n)$ as a Gaussian distribution,
    we need to align it with the perturbations in the semantic space,
    such that the gaussian noise can smooth out the perturbations.
    This actually requires us to regularize the encoder $s(\cdot)$, to meet the certification condition $\hat{R}\leq R$ in Theorem \ref{semanticsmooth}. We formulate this objective as follows:
    %
    \begin{align}
        \label{eq:lrobust}
        \mathcal{L}_{\text{robust}}=  \mathop{\mathbb{E}}_{x,y\sim p(x,y)} \max(0, \hat{R}-R+m),
    \end{align}  
     where $m$ is a hyper-parameter that controls the margin between $\hat{R}$ and $R$.
     As $R$ also has gradients with respect to the parameters of the base classifier $f$,
     this loss also helps optimize the base classifier.

    %
    %
    %

    \textbf{3. IBP Encoder.} The remaining problem is how to get $\hat{R}$ with respect to $x$ and $s(\cdot)$.
    This can be achieved by employing the Interval Bound Propagation (IBP) techniques 
    \citep{weng2018towards,jia2019certified,huang2019achieving}.
    We here employ the method provided by \citet{jia2019certified} to build an IBP encoder $s(\cdot)$,
    and we take  word substitution attacks as an example.
    %
    %
    To be specific, 
    given an IBP encoder $s(\cdot)$, input $x$, and $\hat{x} \in \mathbb{B}_{\text{adv}}(x) = \{\hat{x}: \hat{x}^i \in \mathbb{S}_{\text{adv}}(x^i)\}$,
    we  have:
    \begin{align}
        s_l^i(x) \leq s^i(x) \leq s_u^i(x),
    \end{align}
    where $s^i(x)$ denotes the scalar value of the $i^{th}$ dimension of $s(x)$,
    and  $s_l^i(x)$ and $s_u^i(x)$ are the lowerbound and the upperbound of $s^i(x)$ respectively.

    Thanks to the IBP techniques, here both $s_l^i(x)$ and $s_u^i(x)$ have gradients with respect to $x$ and $\phi$, the parameters of $s(\cdot)$.
    Therefore, it is favorable to  use  $s_l^i(x)$ and $s_u^i(x)$ to
    calculate  $\hat{R}(x)$, as follows:
    \begin{align}
        &\|s(x)-s(\hat{x})\|_2 \leq \hat{R}(x)\\
        &=(\sum_{i} \max(s_u^i(x)-s^i(x),s^i(x)-s_l^i(x))^2)^{\frac{1}{2}},
        \label{eq:hatR}
    \end{align}
    where $\hat{R}(x)$ also has   gradients with respect to $x$ and $\phi$.

    \begin{algorithm}[tb]
        \caption{Training of CISS}
        \label{alg:training}
     \begin{algorithmic}
        \STATE {\bfseries Input:} Data from $p(x,y)$, hyperparameters $\sigma$, $m$, $\gamma$, and the parameters of  Adam.
        \STATE {\bfseries Output:} parameters $\theta$ and $\phi$.
        \REPEAT
        \FOR{random mini-batch  $\sim p(x,y)$}
            \STATE  Compute $\mathcal{L}_{\text{cls}}$ and $\mathcal{L}_{\text{robust}}$ by Eqs. \ref{eq:lcls} and \ref{eq:lrobust}.
            \STATE   Update $\theta$ and $\phi$ by Adam to minimize Eq. \ref{eq:finalobj};
        \ENDFOR
        \UNTIL{the training converges.}
     \end{algorithmic}
     \end{algorithm}

    \textbf{Final Training Objective:}
    Our final training objective is linear combination  of Eqs. \ref{eq:lcls} and \ref{eq:lrobust} with hyperparameter $\gamma$:
    \begin{align}
        \label{eq:finalobj}
        \min_{\theta,\phi}  \mathcal{L}_{\text{cls}} +  \gamma \mathcal{L}_{\text{robust}},
    \end{align}
    where $\theta$ and $\phi$ are the parameters of the encoder $s(\cdot)$ and the base classifier $f(\cdot)$ respectively.

    \setcounter{theorem}{0}

    \begin{remark} 
    \label{remark1}
    (About  $\sigma$)
    In previous randomized smoothing techniques, it is crucial to tune the std $\sigma$ of gaussian noise to an appropriate value:
    a small $\sigma$ will make the final classifier not smoothed enough and thus do harm to an invariant prediction,
    while  a too big $\sigma$ can overly smooth the input and thus impede clean accuracy.
    In contrast, CISS directly minimizes the gap between $R$ and $\hat{R}$, which avoids the hyper-parameter tuning of $\sigma$.
    The encoder $s(\cdot)$ will automatically adapt to  $\sigma$  during the minimization of  $\mathcal{L}_{\text{robust}}$,
     for a better accuracy-robustness trade-off.
    We  support this remark by   ablation on $\sigma$ in Section \ref{sec:ablation} and trade-off  in Section \ref{section trade-off}.
    \end{remark}

     \begin{remark} (About IBP)
         IBP does not fit deep architectures, as the bounds can get looser exponentially with the depth of the model architecture.
    As such many advanced deep neural architectures like Transformers, do not fit IBP well (shown in Table \ref{tab:main result}).
    However, in our framework, we only employ a shallow IBP encoder to map input into a continuous semantic space,
     and thus we benefit from the good mathematical property of IBP while do not suffer from the looseness.
    \end{remark}

    \begin{remark}
    \label{remark3}
    (About the linear combination in Eq.\ref{eq:finalobj})
    We note that $\mathcal{L}_{\text{cls}} +  \gamma \mathcal{L}_{\text{robust}}$ is an upper bound on the expected certification error
    $1 - \mathop{\mathbb{E}}_{x,y\sim p(x,y)} [\mathbbm{1}(R-\hat R>0)]$, as long as $\gamma \geq \frac{1}{m}$ (the proof of which is in the Appendix).
    $\mathcal{L}_{\text{cls}}$ makes the training more smoothed and $\gamma$ should be sufficiently large to effectively minimize the certification error (ablation study on $\gamma$ and $m$ can be found in Section \ref{sec:ablation}).
    %
    
    \end{remark}

    We solve the optimization by Adam \citep{kingma2014adam}.
     The overall training procedure is summarized in Algorithm \ref{alg:training} and illustrated in Fig. \ref{overview}.

    \subsection{Prediction and Robustness Certification of CISS}
    
    %
    %

    \textbf{Prediction.}
    Similar to  \citet{cohen2019certified},
    we employ  Monte Carlo algorithms for our prediction and certification process.
    Given data point $(x,y)$,
    we  build $g$, a hard  prediction version of $f$, to simplify any hypothesis test:
        \begin{align}
            g(x) = \arg\max_{\overline{y}}
          f_{\overline{y}}(s(x)+n),\label{eq:g}
    \end{align}
    with corresponding final prediction c defined as:
    \begin{align}
    c=
    \arg\max_{y'} 
             \mathop{ \mathbb{P}} 
            \big[
            g(x) =y'
            \big].
    \end{align}
    %
    To calculate $c$ with confidence $1-\alpha$, we sample $t$ times and 
    employ a hypothesis test summarized in Algorithm \ref{alg:certification}.

    %

    \textbf{Certification.}
    For the sake of robust certification,
    we employ hypothesis test to ensure our 
    prediction is certifiably robust.
    The justification is
    based on
    Theorem \ref{hardsemanticsmooth} in the Appendix, which consider $g$ as the base classifier instead of $f$.
    %
    %
    The certification procedure is summarized in Algorithm \ref{alg:certification},
    where  function \emph{PvalueBinom} returns the p-value of the two-sided hypothesis test and function  \emph{LowerConfBound} returns the lower bound of estimated Binomial parameter. Detials of \emph{PvalueBinom} and \emph{LowerConfBound} are in appendix \ref{statis_detial}.

     \begin{algorithm}[t]
        \caption{Prediction and Certification of CISS}
        \label{alg:certification}
     \begin{algorithmic}
        
        \STATE {\bfseries Input:} data point $(x,y)$, $\mathbb{B}_{\text{adv}}(x)$, $\sigma$, $f(\cdot)$,  $s(\cdot)$, $t$, and $\alpha$.

        \FUNCTION{\textbf{Prediction}($g, \sigma, x, t; \alpha$)}
            \STATE sample from  $p(n)$ $t$ times to get  $\{n_i\}_1^{t}$
            \STATE For each {$j\in \mathcal{Y}$}, {$\text{cnt}_j$ = $\mathop{\mathbb{E}}_{n\sim \{n_i\}_1^{t}}\mathbbm{1}(g(x)=j)$}
            \STATE $\text{cls}_A$, $\text{cls}_B$ $\leftarrow$ top two indices in $\text{cnt}$
            \STATE {\textbf{If} $\text{PvalueBinom}$($\text{cnt}_A$, $\text{cnt}_A$ + $\text{cnt}_B$, 0.5) $\leq$ $\alpha$, }
            \STATE \textbf{Return} $\text{cls}_A$ with confidence $1-\alpha$
            \STATE \textbf{Else} Abstain from return
        \ENDFUNCTION
        \FUNCTION{\textbf{Certification}($g, \sigma, x, \mathbb{B}_{\text{adv}}(x), t_1, t_2; \alpha$)}
            \STATE sample from  $p(n)$ $t_1+t_2$ times to get  $\{n_i\}_1^{t_1+t_2}$
            \STATE $\text{cls}_A=$  $\arg\max_{j\in \mathcal{Y}} \mathop{\mathbb{E}}_{n\sim \{n_i\}_1^{t_1}}\mathbbm{1}(g(x)=j)$
           \STATE $\text{cnt}_A=$  $\mathop{\mathbb{E}}_{n\sim \{n_i\}_{t_1+1}^{t_1+t_2}}\mathbbm{1}(g(x)=\text{cls}_A)$
            \STATE $\underline{p_{A}}$ = {LowerConfBound}($\text{cnt}_A, t_2, 1-\alpha $)
            
            \STATE \textbf{If} {$\underline{p_{A}} > \frac{1}{2}$} and $\sigma \Phi^{-1}(\underline{p_{A}}) \geq \hat{R}$ ($\hat{R}$ by Eq. \ref{eq:hatR})
                \STATE \textbf{Return} 
                $g(\hat{x})=\text{cls}_A, \forall \hat{x} \in  \mathbb{B}_{\text{adv}}(x)$
                with conf $1-\alpha$
            \STATE \textbf{Else} Abstain from return
        \ENDFUNCTION
     \end{algorithmic}
     \end{algorithm}
    
    \section{Experiments}
    \subsection{Experimental Setting}
    \textbf{Tasks and Datasets.}
    Following previous state-of-the-arts \citep{jia2019certified,ye2020safer}, 
    we examine the certified robustness by text classification tasks,
    and we choose the prevailing YELP\citep{shen2017style} and IMDB\citep{maas2011learning} datasets.
    Also aligned with previous SOTA methods on robust certification,
    we focus on comparing certified robustness against natural language word substitution attacks,
    while we also test and compare the empirical robustness of each method for a more extensive comparison.
    
    \textbf{Baselines.}
    We compare our method with (i) Vanilla BERT \cite{devlin2018bert},  (ii) IBP method  \citep{jia2019certified}, and (iii) SAFER \citep{ye2020safer}, a
    randomized smoothing method based on BERT. As the original IBP method in \citet{jia2019certified} employs shallow text CNNs, we also implement a BERT version for it referred to as IBP-BERT for fair comparisons. We compare with IBP method to show our scalbility to deep architectures, and compare with SAFER to show the benefit of smoothing in the latent space.
    
    \textbf{Certification Setting.}
    We are interested in examining the \textit{certified robust accuracy}, which is defined as the fraction of the test set that is classified correctly with  robust certifications.
    For all randomized smoothing based methods,
    we set $\alpha = 0.001$ to make sure that the certification result for each data point is correct with at least $99.9\%$ confidence. For our method, sampling number is $t_1=50$ and $t_2=30000$ respectively. 
    We use the same word substitution set as in  \citet{jia2019certified}, 
    which is constructed by  the similarity of GloVe word embeddings \cite{pennington2014glove}. 
    The substitution table we use is from \citet{jia2019certified} and is more complicated than that of SAFER \citep{ye2020safer}.
    %
    We do not use any language model constraint on the generated adversarial examples, and there is no limit on the number of substitutions per input.
    
    \begin{figure}[t]
        \centering 
        \vspace{-0pt} 
        \includegraphics[width=1\linewidth]{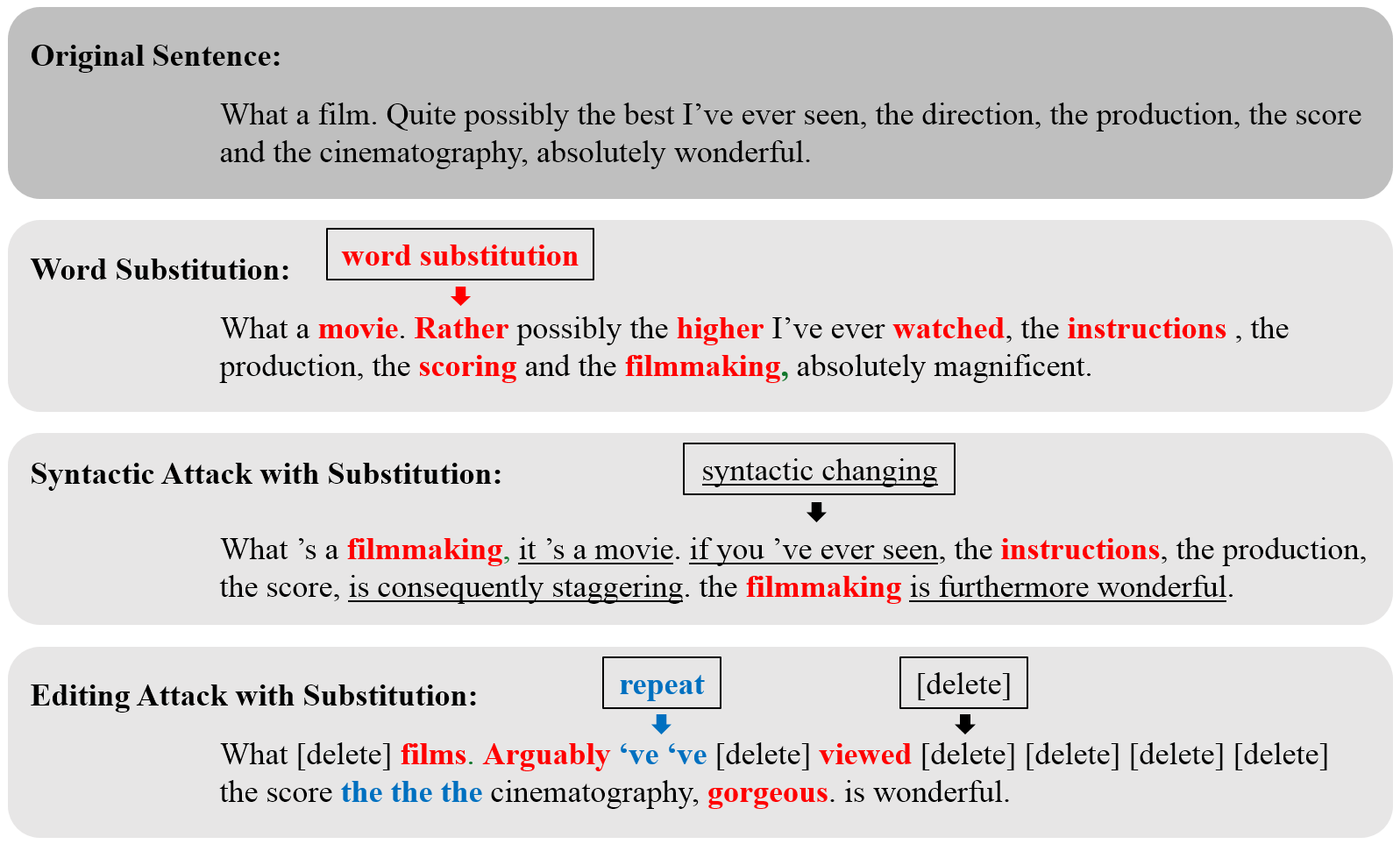}
        \vspace{-6mm}
        \caption{An illustration of the attacks we employed, using a random example from IMDB.}
       \label{fig:attacks}
        \vspace{-5mm}
    \end{figure}

    \textbf{Empirical Setting Against Word Substitutions.}
    In this setting we aim to examine the empirical robustness of each  method against a real  attack algorithm.
    This also serves as a sanity check of our  certification process.
    We  employ ASCC \citep{dong2021towards}, 
    an empirical word substitution attack 
    that finds the worst-case perturbation inside the convex hull composed of substitutions.

    \textbf{Empirical Setting Against Word Substitutions Combined With Unseen Attacks.}
    In this setting we aim to examine whether the proposed framework is robust against some unseen natural language attacks,
    when we only see word substitution attacks during training.
    %
    (1) HiddenKiller \citep{qi2021hidden}, a syntactic-based attack, It paraphrases the original input sentence by altering its parser tree; word substitution and irrelevant sentences are added to strengthen this attack. 
    (2) Editing attack, which is inspired by \cite{liang2017deep}, not only substitutes words but also change the token positions and sentence structures, by adding repetitive words,
    substituting words, and deleting random words.
    This two  attacks are unseen during the training phrase of our model and we further enhanced them by combining them  with word substitution attacks.
    We give some qualitative examples of these attacks in Fig. \ref{fig:attacks}.
    
    
    \textbf{Implementation Details.} 
    In our experiments, we employ IBP-based CNN as our encoder $s$.
    For the base classifier, we use BERT (base-uncased) \citep{devlin2018bert}.
    As BERT takes natural language as input, we additionally add a CNN decoder before BERT.
    We note that our framework can be also applied to  other NLP architectures and if the base classifier is not a pre-trained language model then the decoder might not be necessary.
    %
    For hyper-parameters, we set $\sigma=1$, $\gamma = 4.0$, and margin $m = 1.0$ (ablation on hyper-parameters in section \ref{sec:ablation}). These parameters are tuned to achieve the best certified robustness as shown in \ref{sec:ablation}. During training, we first use loss $\mathcal{L}_{\text{cls}}$ to optimize the model to convergence, and then add loss $\mathcal{L}_{\text{robust}}$ for training. Warm-up is used on $\gamma$ during optimization. During training, we sample only 1 time from the Gaussian to perform smoothing.
    %
    For ASCC attack, we run for $10$ iterations  to find the worst-case attack, and then
    discretize the  attack into textual adversarial examples.
    In editing attack, we use a editing distance of 10 and 50 on YELP and IMDB, respectively. Our code is available at https://github.com/zhao-ht/ConvexCertify.
    %

    \subsection{Main Results} 
    
    In this subsection, 
    we examine the certified robustness against word substitutions,
    and compare our method with state-of-the-arts.
    As shown in Table \ref{tab:main result},
    our model  outperforms  baselines  on both IMDB and YELP with significant margin.
    Specifically,
    we achieve 90.58\% certified robustness on YELP,
    and 76.45\% certified robustness on IMDB,
    surpassing the runner-up by 6.8\% and 7.2\% respectively.
    The significant margin strongly validates the certified robustness of the proposed CISS framework.
   
   \begin{table}[t]\centering
    
        \begin{tabular}{lcc}
        \hline
        \hline
                               Method  & YELP  & IMDB  \\ \hline
        IBP \citep{jia2019certified} & 83.81 & 68.60 \\
        IBP-BERT                     &   N.A.   & N.A.     \\
        
        SAFER \citep{ye2020safer}    & 80.63 & 69.20 \\

        CISS (Ours)                  & \textbf{90.58} & \textbf{76.45} \\
        \hline
        \hline
        \end{tabular}
    
    \caption{Certified robust accuracy ($\%$) against word substitutions on  YELP and IMDB. All compared methods use the same word substitution table from \cite{jia2019certified} for a fair comparison.
    }
    \label{tab:main result}
    \end{table}

    \begin{table}[t]\centering
        \begin{tabular}{lcc}
        \hline
        \hline
        Method                       & YELP           & IMDB           \\ \hline
        IBP \citep{jia2019certified} & 84.45          & 74.80           \\
        Vanilla BERT                 & 50.09          & 5.68           \\
        SAFER \citep{ye2020safer}    & 80.63          & 69.20           \\
        CISS (Ours)                  & \textbf{91.72} & \textbf{78.29} \\ \hline
        \hline
        \end{tabular}
    
    \caption{Empirical robustness ($\%$) of each method on the YELP and IMDB, under the ASCC word substitution attack (using the same substitution table as in Table  \ref{tab:main result}).}
    \label{ascc result}
    \vspace{-5mm} 
    \end{table}
    
\begin{figure}[t]

        \hspace{9mm} 
        \includegraphics[width=0.8\linewidth]{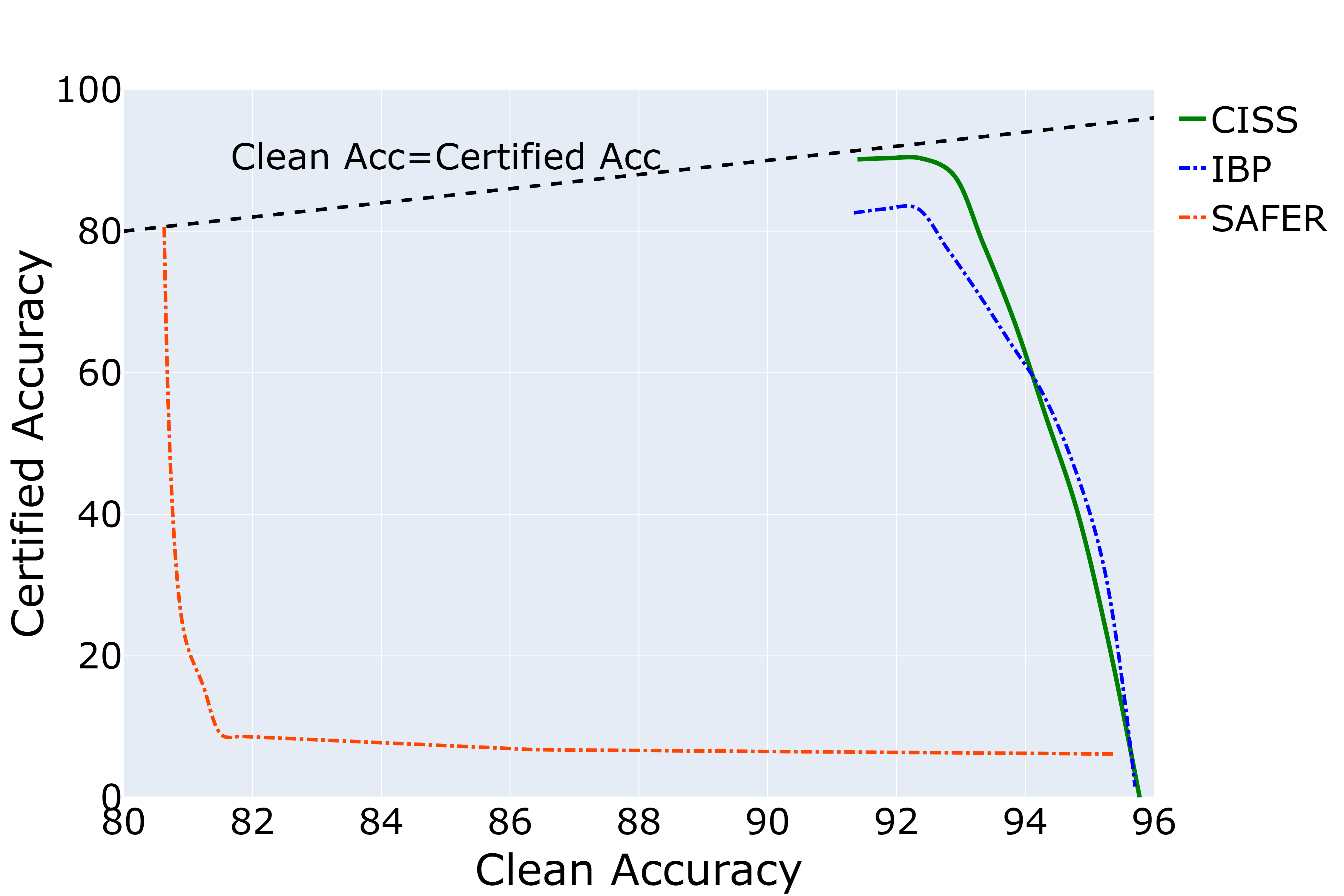}
        
        \caption{The trade-off curve between robustness and clean accuracy of compared methods. The black dashed line is 
        the the upper bound of such a trade-off, as a model's robust accuracy is always lower than its clean accuracy.}
        \label{fig:curve}
        
        \vspace{-5mm}
    \end{figure}

    Though IBP can achieve good certified robustness by using only shallow architectures like CNN, it hardly scales to deep models.
    For example, we test the certified robustness of IBP with  BERT and the result is N.A.; this is because the interval bound grows exponentially with the depth of the architecture and finally becomes too loose to be used.
    SAFER smoothes in the input textual space
    but it relies on customized  noise distribution over substitution words.
    To achieve robustness, its noise distribution over words forms a graph where many semantically unrelated words are forced to be connected, which impedes the discriminative power of a model and degrades its clean accuracy severely.
    See Fig. \ref{fig:curve}, where SAFER needs to sacrifice the clean accuracy from 96\% down to 82\% to increase some robustness.
    %
    On the contrary,
    CISS smoothes in the latent semantic space and does not suffer from this problem.
    
    

    \subsection{Empirical Robustness Against Word Substitutions}
    In this section we employ a real word substitution attack , ASCC \citep{dong2021towards}, to
    examine the robustness of each compared method.
    The empirical robust accuracy also help us sanity check our certification procedure.
    The result is shown in Table \ref{ascc result}.
    As we can see, the ASCC attack is quite strong: it  degrades the accuracy of a vanilla BERT to $5.68\%$ (from more than $92\%$),
    and the empirical robust accuracy of compared methods  are just slightly higher than their corresponding certified robust accuracy.
    %
    %
    Nonetheless, CISS still outperforms all baselines with significant margins; \emph{i.e.}, CISS surpasses the runner-up by 7.3$\%$ on YELP and 3.5$\%$ on IMDB, respectively. 
    %

    \subsection{Robustness Against 
    Word Substitutions Combined With  Unseen Natural Language Attacks}
    As proposed in our causal graph \ref{causal_graph},
    attacks leverage the spurious correlations from $X\leftarrow N \rightarrow Y$ to fool a model, by manipulating $N$.
    There exist many different confounders manipulatable by different kind of attacks;
    \emph{e.g.}, strict professional movie reviewer might prefer specific kind of syntactic structures.
    However, in training, we may not be able to enumerate all possible attacks to removing all possible manipulatable confounders.
    Here we conduct experiments and 
    examine whether the proposed framework is still robust
    when there exist unseen attacks  during testing.
    Specifically, we employ two sentence-level attacks that are both unseen during training, Syntactic-based attack \citep{qi2021hidden} and Editing attack \citep{liang2017deep} ,
    and further strength them with word substitution attacks,
    to examine whether the proposed method is still robust under such a challenge setting.
    As shown in Table \ref{sentence result}, CISS still remains certain level of robustness, and still outperforms state-of-the-arts by around $7\%$.
    This owes to our smoothing in the semantic space,
    as it can remove some other latent  confounder effects manipulatable by certain unseen attacks, and thus provide more generalized NLP robustness,
    while directly smoothing in the input space may not.

    \begin{table}[t]
   
    \centering
        \resizebox{\columnwidth}{11mm}{
            \begin{tabular}{l|cc|cc}
            \hline
            \hline
            Method       & \multicolumn{2}{c|}{YELP}     & \multicolumn{2}{c}{IMDB} \\ \hline
            Attack Method & Syntactic     & Editing       & Syntactic     & Editing  \\ \hline
            IBP          & 73.7          & 75.3          & 67.7           & 67.1     \\
            SAFER        & 67.1          & 69.0           & 66.4           & 66.2      \\
            CISS (Ours)  & \textbf{80.7} & \textbf{83.1} & \textbf{74.4}  & \textbf{74.3}     \\ 
            \hline
            \hline
            \end{tabular}
            }
        
        \caption{The certified accuracy against word substitution of our method and baselines on the YELP and IMDB datasets with sentence level confounders. Our model outperforms baselines under both syntactic and editing sentence level perturbations.}
        \label{sentence result}
         \vspace{-5mm}
        \end{table}

    \subsection{Accuracy-Robustness Trade-Off} \label{section trade-off}
    
    Here we show the trade-off curves between certified robustness and clean accuracy of each method.
    For CISS, the trade-off is achieved by tuning $\gamma$.
    For SAFER, it is achieved by adjusting the scale of noise.
    For  IBP, it is achieved by tuning the coefficient of training loss.
    
    Figure \ref{fig:curve} clearly shows that CISS achieves a better trade-off between robustness and clean accuracy compared to baselines.
     CISS achieves higher certified robust accuracy without sacrificing too much clean accuracy,
     and our curve is very closer to the theoretical upper bound, \emph{i.e.}, the dashed black line representing clean accuracy = certified robust accuracy (as robust accuracy is always lower than the clean accuracy).
     This owes to that our encoder automatically adapts to our gaussian std $\sigma$ towards the best  trade-off curve of CISS, as we mentioned in Remark \ref{remark1}.


    \subsection{Ablation Study and Hyper-parameter Sensitivity}
    \label{sec:ablation}
    In this section, we do the ablation study by training CISS using different values of $\gamma$, $m$, and $\sigma$,
    and see how each hyper-parameter affects the performance. In this section, for the sake of efficiency, we only sample $t_2=300$ times in the second step of the certification, and set $\alpha$ to 0.05 so that we can obtain results comparable with Tabel \ref{tab:main result}. See the Appendix \ref{alpha300l} for details.
    
    First, Table \ref{ablation YELP} and Table \ref{ablation IMDB} illustrate the certified robustness of CISS on YELP and IMDB using different values of $\gamma$ and $m$. 
    We observe that low $\gamma$ like 0.25 and $m$ like -0.5 can degrade robustness;
    When $\gamma$ and $m$ are sufficiently large,
    the final performance is not sensitive to small changes of $\gamma$ and $m$.
    %
    %
    %
    The above result  supports the claim 
    we made in Remark \ref{remark3}:
    our training objective $\mathcal{L}_{\text{cls}} +  \gamma \mathcal{L}_{\text{robust}}$ is an upper bound on the certification error  $1 - \mathop{\mathbb{E}}_{x,y\sim p(x,y)} [\mathbbm{1}(R-\hat R>0)]$, when $\gamma * m \ge 1$.
    Too small $\gamma$ and $m$ cannot ensure that our training objective is an upper bound and thus cannot support effective minimization of the certification error;
    when $\gamma * m \ge 1$ is sufficiently large, our training objective  is already an upper bound on the certification error and thus the performance is not sensitive to small changes of $\gamma$ or $m$. 
    %
    
    \begin{table}[t]
    \resizebox{\columnwidth}{11mm}{
        \begin{tabular}{lc|lc}
        \hline
        \hline
        When $m=1$    & Robustness & When $\gamma=4$    & Robustness \\ \hline
        $\gamma = 0.25$ & 73.38            & $m=-0.5$  & 81.70            \\
        $\gamma = 2$    & 89.72           & $m= 0 $   & 89.65           \\
        $\gamma = 4$    & \textbf{90.47}           &  $m= 1 $ &  \textbf{90.47}            \\
        $\gamma = 8$    & 90.15            & $m= 2$    & 90.33           \\ \hline
        \hline
        \end{tabular}
    }
    \caption{Certified robust accuracy ($\%$) of our method on YELP using different values of $\gamma$ and $m$.}
    \label{ablation YELP}
    \end{table}
    
    \begin{table}[]
    \resizebox{\columnwidth}{11mm}{
        \begin{tabular}{lc|lc}
        \hline
        \hline
        When $m=1$      & Robustness & When $\gamma=4$     & Robustness \\ \hline
        $\gamma = 0.25$ & 48.13         & $m=-0.5$  & 66.40            \\
        $\gamma = 2 $  & 71.20           & $m= 0 $   & 70.26           \\
        \textbf{$\gamma = 4$}    &   \textbf{75.25}          &  \textbf{$m= 1 $} &  \textbf{75.25}            \\
        $\gamma = 8 $    &  74.37     & $m= 2$    &    75.20       \\ \hline
        \hline
        \end{tabular}
    }
    \caption{Certified robust accuracy ($\%$) of our method on IMDB using different values of $\gamma$ and $m$.}
    \label{ablation IMDB}
    \vspace{-1mm}
    \end{table}

    We also conduct experiments to 
    examine how the Gaussian std $\sigma$ 
    affect the performance of CISS, 
    in order to support our Remark \ref{remark1}.
    Table \ref{ablation sigma} shows how different values of $\sigma$ affect the performance of CISS.
    As demonstrated, 
    changing the  value of $\sigma$ only affect the final performance to an ignorable extent;
    all the certified robustness showed in Table \ref{ablation sigma} are strong and  surpass state-of-the-arts.
    Therefore, it can be argued that  CISS does not rely on tedious tuning of $\sigma$ to achieve a good performance, while previous randomized smoothing methods rely on the tuning of $\sigma$ heavily.

    \begin{table}[t]
    \resizebox{\columnwidth}{6mm}{
        \begin{tabular}{lllll}
        \hline
        \hline
        Gaussian  Std $\sigma$    & 0.5 & 1    & 2 & 4 \\ 
        \hline
        Certified Robustness & 90.30 & 90.47 & 90.43 & 90.14 \\
        \hline
        \hline
        \end{tabular}
    }
    
    \caption{Certified robust accuracy  ($\%$) of our method on YELP using different values of Gaussian std $\sigma$.}
    \vspace{-5mm}
    \label{ablation sigma}
    \end{table}

\section{Related work}
\label{related work}
\textbf{Causality and Adversarial Robustness.}
Graphical causal inference \citep{pearl2009causal,peters2017elements}
aims at discovering the causal structure, calculating the causal effect of interventions,
and answering counterfactual questions.
’s interest in causality has significantly increased in recent years
In recent years,
it has drawn increasing attention from the machine learning community \cite{scholkopf2019causality,scholkopf2021toward},
\emph{e.g.}, in few-shot learning \cite{teshima2020few}\cite{yue2020interventional},
long-tail classification \cite{tang2020long},
and generative modeling \cite{sauer2021counterfactual}.
Causal inference has an instinct for modeling
distribution change and adversarial robustness,
\emph{e.g.},  \citet{zhang2020causal}
\citet{yang2019causal}
\citet{mitrovic2020representation} \citet{tang2021adversarial}.
This work differs in that we 
propose a causal perspective to understand 
 randomized smoothing, and achieve certified robustness by modeling causal effects.

\vspace{0mm}
\textbf{NLP Attacks and Empirical Defenses.}
Adversarial attacks \citep{szegedy2013intriguing,goodfellow2014explaining,papernot2016limitations,kurakin2016adversarial},
are maliciously generated to fool
DNNs while keeping innocuous to humans.
   %
   %
   In NLP, 
     attacks algorithms include char-level modifications 
   \citep{hosseini2017deceiving,ebrahimi2017hotflip,belinkov2017synthetic, gao2018black,eger2019text,pruthi2019combating},
   sequence-level manipulations 
    \citep{iyyer2018adversarial,ribeiro2018semantically,jia2017adversarial,zhao2017generating,qi2021hidden},
   %
   and  adversarial word substitutions  
    \citep{alzantot2018generating,ren2019generating,jin2019bert,dong2021towards,zang2020sememe}.
Adversarial training 
 \citep{madry2017towards,athalye2018obfuscated, miyato2016adversarial, ebrahimi2017hotflip, dong2021towards,dong2021should}
 augments training by adversarial examples and
 is currently the most effective empirical defense in NLP defense.
   
\textbf{Certified Robustness and Randomized Smoothing.} 
This line of work aims at provable robustness with theoretical guarantees.
%
%
This field can be sorted into deterministic methods and randomized smoothing methods. Deterministic certification includes Dual Approach \citep{dvijotham2018training,Dvijotham2018ADA}, 
Interval Bound Propagation (IBP) \citep{wong2018provable,jia2019certified,huang2019achieving} and Linear Relaxation methods \citep{zhang2018crown, zhang2019stable, shi2020robustness}. IBP methods certify the robustness by propagating
interval bounds to bound the output, but  are often too computationally expensive. %
In contrast, randomized smoothing \citep{lecuyer2019certified,cohen2019certified}  are more applicable  to  large models \cite{Zhai2020MACERAA}.
Due to the discrete NLP input space, \citet{zeng2021certified}  masks  tokens randomly, and SAFER \citep{ye2020safer} constructs stochastic ensembles by random substitutions.
%
Our method differs in that we smooth in the latent space, which makes our framework applicable to more general NLP attacks and frees us from constructing complicated attack-customized noise distributions.  

\section{Discussion and Conclusion}
In this paper, we provide a novel causal perspective  to understand model vulnerability,
and propose a novel framework CISS to achieve certified NLP robustness.
CISS consistently surpasses state-of-the-arts with significant margins.
In future work, we plan to explore a more generalized framework towards robustness in both CV and NLP.

\section*{Acknowledgements} 

This work is supported by the Singapore Ministry of Education (MOE) Academic Research Fund
(AcRF) Tier 1 (RS21/20) and Tier 2.

\bibliography{submission}

\begin{thebibliography}{69}
\providecommand{\natexlab}[1]{#1}
\providecommand{\url}[1]{\texttt{#1}}
\expandafter\ifx\csname urlstyle\endcsname\relax
  \providecommand{\doi}[1]{doi: #1}\else
  \providecommand{\doi}{doi: \begingroup \urlstyle{rm}\Url}\fi

\bibitem[Aldrich(1989)]{aldrich1989autonomy}
Aldrich, J.
\newblock Autonomy.
\newblock \emph{Oxford Economic Papers}, 41\penalty0 (1):\penalty0 15--34,
  1989.

\bibitem[Alzantot et~al.(2018)Alzantot, Sharma, Elgohary, Ho, Srivastava, and
  Chang]{alzantot2018generating}
Alzantot, M., Sharma, Y., Elgohary, A., Ho, B.-J., Srivastava, M., and Chang,
  K.-W.
\newblock Generating natural language adversarial examples.
\newblock In \emph{EMNLP}, 2018.

\bibitem[Athalye et~al.(2018)Athalye, Carlini, and
  Wagner]{athalye2018obfuscated}
Athalye, A., Carlini, N., and Wagner, D.
\newblock Obfuscated gradients give a false sense of security: Circumventing
  defenses to adversarial examples.
\newblock In \emph{ICML}, 2018.

\bibitem[Belinkov \& Bisk(2018)Belinkov and Bisk]{belinkov2017synthetic}
Belinkov, Y. and Bisk, Y.
\newblock Synthetic and natural noise both break neural machine translation.
\newblock In \emph{ICLR}, 2018.

\bibitem[Cohen et~al.(2019)Cohen, Rosenfeld, and Kolter]{cohen2019certified}
Cohen, J., Rosenfeld, E., and Kolter, Z.
\newblock Certified adversarial robustness via randomized smoothing.
\newblock In \emph{International Conference on Machine Learning}. PMLR, 2019.

\bibitem[Devlin et~al.(2019)Devlin, Chang, Lee, and Toutanova]{devlin2018bert}
Devlin, J., Chang, M.-W., Lee, K., and Toutanova, K.
\newblock Bert: Pre-training of deep bidirectional transformers for language
  understanding.
\newblock In \emph{NAACL}, 2019.

\bibitem[Dong et~al.(2021{\natexlab{a}})Dong, Luu, Ji, and
  Liu]{dong2021towards}
Dong, X., Luu, A.~T., Ji, R., and Liu, H.
\newblock Towards robustness against natural language word substitutions.
\newblock In \emph{ICLR}, 2021{\natexlab{a}}.

\bibitem[Dong et~al.(2021{\natexlab{b}})Dong, Luu, Lin, Yan, and
  Zhang]{dong2021should}
Dong, X., Luu, A.~T., Lin, M., Yan, S., and Zhang, H.
\newblock How should pre-trained language models be fine-tuned towards
  adversarial robustness?
\newblock In \emph{NeurIPS}, 2021{\natexlab{b}}.

\bibitem[Dvijotham et~al.(2018{\natexlab{a}})Dvijotham, Gowal, Stanforth,
  Arandjelovic, O'Donoghue, Uesato, and Kohli]{dvijotham2018training}
Dvijotham, K., Gowal, S., Stanforth, R., Arandjelovic, R., O'Donoghue, B.,
  Uesato, J., and Kohli, P.
\newblock Training verified learners with learned verifiers.
\newblock \emph{arXiv preprint arXiv:1805.10265}, 2018{\natexlab{a}}.

\bibitem[Dvijotham et~al.(2018{\natexlab{b}})Dvijotham, Stanforth, Gowal, Mann,
  and Kohli]{Dvijotham2018ADA}
Dvijotham, K., Stanforth, R., Gowal, S., Mann, T.~A., and Kohli, P.
\newblock A dual approach to scalable verification of deep networks.
\newblock \emph{ArXiv}, abs/1803.06567, 2018{\natexlab{b}}.

\bibitem[Ebrahimi et~al.(2018)Ebrahimi, Rao, Lowd, and
  Dou]{ebrahimi2017hotflip}
Ebrahimi, J., Rao, A., Lowd, D., and Dou, D.
\newblock Hotflip: White-box adversarial examples for text classification.
\newblock In \emph{ACL}, 2018.

\bibitem[Eger et~al.(2019)Eger, {\c{S}}ahin, R{\"u}ckl{\'e}, Lee, Schulz,
  Mesgar, Swarnkar, Simpson, and Gurevych]{eger2019text}
Eger, S., {\c{S}}ahin, G.~G., R{\"u}ckl{\'e}, A., Lee, J.-U., Schulz, C.,
  Mesgar, M., Swarnkar, K., Simpson, E., and Gurevych, I.
\newblock Text processing like humans do: Visually attacking and shielding nlp
  systems.
\newblock In \emph{NAACL}, 2019.

\bibitem[Gao et~al.(2018)Gao, Lanchantin, Soffa, and Qi]{gao2018black}
Gao, J., Lanchantin, J., Soffa, M.~L., and Qi, Y.
\newblock Black-box generation of adversarial text sequences to evade deep
  learning classifiers.
\newblock In \emph{SPW}. IEEE, 2018.

\bibitem[Geirhos et~al.(2020)Geirhos, Jacobsen, Michaelis, Zemel, Brendel,
  Bethge, and Wichmann]{geirhos2020shortcut}
Geirhos, R., Jacobsen, J.-H., Michaelis, C., Zemel, R., Brendel, W., Bethge,
  M., and Wichmann, F.~A.
\newblock Shortcut learning in deep neural networks.
\newblock \emph{Nature Machine Intelligence}, 2\penalty0 (11):\penalty0
  665--673, 2020.

\bibitem[Goodfellow et~al.(2016)Goodfellow, Bengio, Courville, and
  Bengio]{goodfellow2016deep}
Goodfellow, I., Bengio, Y., Courville, A., and Bengio, Y.
\newblock \emph{Deep learning}, volume~1.
\newblock MIT Press, 2016.

\bibitem[Goodfellow et~al.(2015)Goodfellow, Shlens, and
  Szegedy]{goodfellow2014explaining}
Goodfellow, I.~J., Shlens, J., and Szegedy, C.
\newblock Explaining and harnessing adversarial examples.
\newblock In \emph{ICLR}, 2015.

\bibitem[Gopnik et~al.(2004)Gopnik, Glymour, Sobel, Schulz, Kushnir, and
  Danks]{gopnik2004theory}
Gopnik, A., Glymour, C., Sobel, D.~M., Schulz, L.~E., Kushnir, T., and Danks,
  D.
\newblock A theory of causal learning in children: causal maps and bayes nets.
\newblock \emph{Psychological review}, 111\penalty0 (1):\penalty0 3, 2004.

\bibitem[Hinton et~al.(2012)Hinton, Deng, Yu, Dahl, Mohamed, Jaitly, Senior,
  Vanhoucke, Nguyen, Sainath, et~al.]{hinton2012deep}
Hinton, G., Deng, L., Yu, D., Dahl, G.~E., Mohamed, A.-r., Jaitly, N., Senior,
  A., Vanhoucke, V., Nguyen, P., Sainath, T.~N., et~al.
\newblock Deep neural networks for acoustic modeling in speech recognition: The
  shared views of four research groups.
\newblock \emph{IEEE Signal processing magazine}, 2012.

\bibitem[Hoover(2008)]{hoover2008causality}
Hoover, K.~D.
\newblock Causality in economics and econometrics.
\newblock \emph{The new Palgrave dictionary of economics}, 2, 2008.

\bibitem[Hosseini et~al.(2017)Hosseini, Kannan, Zhang, and
  Poovendran]{hosseini2017deceiving}
Hosseini, H., Kannan, S., Zhang, B., and Poovendran, R.
\newblock Deceiving google's perspective api built for detecting toxic
  comments.
\newblock \emph{arXiv preprint arXiv:1702.08138}, 2017.

\bibitem[Huang et~al.(2019)Huang, Stanforth, Welbl, Dyer, Yogatama, Gowal,
  Dvijotham, and Kohli]{huang2019achieving}
Huang, P.-S., Stanforth, R., Welbl, J., Dyer, C., Yogatama, D., Gowal, S.,
  Dvijotham, K., and Kohli, P.
\newblock Achieving verified robustness to symbol substitutions via interval
  bound propagation.
\newblock In \emph{EMNLP}, 2019.

\bibitem[Ilyas et~al.(2019)Ilyas, Santurkar, Tsipras, Engstrom, Tran, and
  Madry]{ilyas2019adversarial}
Ilyas, A., Santurkar, S., Tsipras, D., Engstrom, L., Tran, B., and Madry, A.
\newblock Adversarial examples are not bugs, they are features.
\newblock \emph{arXiv preprint arXiv:1905.02175}, 2019.

\bibitem[Iyyer et~al.(2018)Iyyer, Wieting, Gimpel, and
  Zettlemoyer]{iyyer2018adversarial}
Iyyer, M., Wieting, J., Gimpel, K., and Zettlemoyer, L.
\newblock Adversarial example generation with syntactically controlled
  paraphrase networks.
\newblock In \emph{NAACL}, 2018.

\bibitem[Jia \& Liang(2017)Jia and Liang]{jia2017adversarial}
Jia, R. and Liang, P.
\newblock Adversarial examples for evaluating reading comprehension systems.
\newblock In \emph{EMNLP}, 2017.

\bibitem[Jia et~al.(2019)Jia, Raghunathan, G{\"o}ksel, and
  Liang]{jia2019certified}
Jia, R., Raghunathan, A., G{\"o}ksel, K., and Liang, P.
\newblock Certified robustness to adversarial word substitutions.
\newblock In \emph{EMNLP}, 2019.

\bibitem[Jin et~al.(2020)Jin, Jin, Zhou, and Szolovits]{jin2019bert}
Jin, D., Jin, Z., Zhou, J.~T., and Szolovits, P.
\newblock Is bert really robust? natural language attack on text classification
  and entailment.
\newblock \emph{AAAI}, 2020.

\bibitem[Kingma \& Ba(2015)Kingma and Ba]{kingma2014adam}
Kingma, D.~P. and Ba, J.
\newblock Adam: A method for stochastic optimization.
\newblock In \emph{ICLR}, 2015.

\bibitem[Krizhevsky et~al.(2012)Krizhevsky, Sutskever, and
  Hinton]{krizhevsky2012imagenet}
Krizhevsky, A., Sutskever, I., and Hinton, G.~E.
\newblock Imagenet classification with deep convolutional neural networks.
\newblock In \emph{NeurIPS}, 2012.

\bibitem[Kurakin et~al.(2016)Kurakin, Goodfellow, and
  Bengio]{kurakin2016adversarial}
Kurakin, A., Goodfellow, I., and Bengio, S.
\newblock Adversarial examples in the physical world, 2016.

\bibitem[Lecuyer et~al.(2019)Lecuyer, Atlidakis, Geambasu, Hsu, and
  Jana]{lecuyer2019certified}
Lecuyer, M., Atlidakis, V., Geambasu, R., Hsu, D., and Jana, S.
\newblock Certified robustness to adversarial examples with differential
  privacy.
\newblock In \emph{2019 IEEE Symposium on Security and Privacy (SP)}, pp.\
  656--672. IEEE, 2019.

\bibitem[Levenshtein et~al.(1966)]{levenshtein1966binary}
Levenshtein, V.~I. et~al.
\newblock Binary codes capable of correcting deletions, insertions, and
  reversals.
\newblock In \emph{Soviet physics doklady}. Soviet Union, 1966.

\bibitem[Liang et~al.(2018)Liang, Li, Su, Bian, Li, and Shi]{liang2017deep}
Liang, B., Li, H., Su, M., Bian, P., Li, X., and Shi, W.
\newblock Deep text classification can be fooled.
\newblock In \emph{IJCAI}, 2018.

\bibitem[Maas et~al.(2011)Maas, Daly, Pham, Huang, Ng, and
  Potts]{maas2011learning}
Maas, A., Daly, R.~E., Pham, P.~T., Huang, D., Ng, A.~Y., and Potts, C.
\newblock Learning word vectors for sentiment analysis.
\newblock In \emph{Proceedings of the 49th annual meeting of the association
  for computational linguistics: Human language technologies}, pp.\  142--150,
  2011.

\bibitem[Madry et~al.(2018)Madry, Makelov, Schmidt, Tsipras, and
  Vladu]{madry2017towards}
Madry, A., Makelov, A., Schmidt, L., Tsipras, D., and Vladu, A.
\newblock Towards deep learning models resistant to adversarial attacks.
\newblock In \emph{ICLR}, 2018.

\bibitem[Mitrovic et~al.(2020)Mitrovic, McWilliams, Walker, Buesing, and
  Blundell]{mitrovic2020representation}
Mitrovic, J., McWilliams, B., Walker, J., Buesing, L., and Blundell, C.
\newblock Representation learning via invariant causal mechanisms.
\newblock \emph{arXiv preprint arXiv:2010.07922}, 2020.

\bibitem[Miyato et~al.(2017)Miyato, Dai, and Goodfellow]{miyato2016adversarial}
Miyato, T., Dai, A.~M., and Goodfellow, I.
\newblock Adversarial training methods for semi-supervised text classification.
\newblock In \emph{ICLR}, 2017.

\bibitem[Papernot et~al.(2016)Papernot, McDaniel, Jha, Fredrikson, Celik, and
  Swami]{papernot2016limitations}
Papernot, N., McDaniel, P., Jha, S., Fredrikson, M., Celik, Z.~B., and Swami,
  A.
\newblock The limitations of deep learning in adversarial settings.
\newblock In \emph{EuroS\&P}. IEEE, 2016.

\bibitem[Pearl(2009)]{pearl2009causal}
Pearl, J.
\newblock Causal inference in statistics: An overview.
\newblock \emph{Statistics surveys}, 3:\penalty0 96--146, 2009.

\bibitem[Pennington et~al.(2014)Pennington, Socher, and
  Manning]{pennington2014glove}
Pennington, J., Socher, R., and Manning, C.~D.
\newblock Glove: Global vectors for word representation.
\newblock In \emph{EMNLP}, pp.\  1532--1543, 2014.

\bibitem[Peters et~al.(2017)Peters, Janzing, and
  Sch{\"o}lkopf]{peters2017elements}
Peters, J., Janzing, D., and Sch{\"o}lkopf, B.
\newblock \emph{Elements of causal inference: foundations and learning
  algorithms}.
\newblock The MIT Press, 2017.

\bibitem[Pruthi et~al.(2019)Pruthi, Dhingra, and Lipton]{pruthi2019combating}
Pruthi, D., Dhingra, B., and Lipton, Z.~C.
\newblock Combating adversarial misspellings with robust word recognition.
\newblock In \emph{ACL}, 2019.

\bibitem[Qi et~al.(2021)Qi, Li, Chen, Zhang, Liu, Wang, and Sun]{qi2021hidden}
Qi, F., Li, M., Chen, Y., Zhang, Z., Liu, Z., Wang, Y., and Sun, M.
\newblock Hidden killer: Invisible textual backdoor attacks with syntactic
  trigger.
\newblock In \emph{ACL}, 2021.

\bibitem[Ren et~al.(2015)Ren, He, Girshick, and Sun]{ren2015faster}
Ren, S., He, K., Girshick, R., and Sun, J.
\newblock Faster r-cnn: Towards real-time object detection with region proposal
  networks.
\newblock In \emph{NeurIPS}, 2015.

\bibitem[Ren et~al.(2019)Ren, Deng, He, and Che]{ren2019generating}
Ren, S., Deng, Y., He, K., and Che, W.
\newblock Generating natural language adversarial examples through probability
  weighted word saliency.
\newblock In \emph{ACL}, 2019.

\bibitem[Ribeiro et~al.(2018)Ribeiro, Singh, and
  Guestrin]{ribeiro2018semantically}
Ribeiro, M.~T., Singh, S., and Guestrin, C.
\newblock Semantically equivalent adversarial rules for debugging nlp models.
\newblock In \emph{ACL}, 2018.

\bibitem[Sauer \& Geiger(2021)Sauer and Geiger]{sauer2021counterfactual}
Sauer, A. and Geiger, A.
\newblock Counterfactual generative networks.
\newblock \emph{arXiv preprint arXiv:2101.06046}, 2021.

\bibitem[Sch{\"o}lkopf(2019)]{scholkopf2019causality}
Sch{\"o}lkopf, B.
\newblock Causality for machine learning.
\newblock \emph{arXiv preprint arXiv:1911.10500}, 2019.

\bibitem[Sch{\"o}lkopf et~al.(2021)Sch{\"o}lkopf, Locatello, Bauer, Ke,
  Kalchbrenner, Goyal, and Bengio]{scholkopf2021toward}
Sch{\"o}lkopf, B., Locatello, F., Bauer, S., Ke, N.~R., Kalchbrenner, N.,
  Goyal, A., and Bengio, Y.
\newblock Toward causal representation learning.
\newblock \emph{Proceedings of the IEEE}, 109\penalty0 (5):\penalty0 612--634,
  2021.

\bibitem[Shen et~al.(2017)Shen, Lei, Barzilay, and Jaakkola]{shen2017style}
Shen, T., Lei, T., Barzilay, R., and Jaakkola, T.
\newblock Style transfer from non-parallel text by cross-alignment.
\newblock \emph{arXiv preprint arXiv:1705.09655}, 2017.

\bibitem[Shi et~al.(2020)Shi, Zhang, Chang, Huang, and
  Hsieh]{shi2020robustness}
Shi, Z., Zhang, H., Chang, K.-W., Huang, M., and Hsieh, C.-J.
\newblock Robustness verification for transformers, 2020.

\bibitem[Sutskever et~al.(2014)Sutskever, Vinyals, and
  Le]{sutskever2014sequence}
Sutskever, I., Vinyals, O., and Le, Q.~V.
\newblock Sequence to sequence learning with neural networks.
\newblock In \emph{NeurIPS}, 2014.

\bibitem[Szegedy et~al.(2013)Szegedy, Zaremba, Sutskever, Bruna, Erhan,
  Goodfellow, and Fergus]{szegedy2013intriguing}
Szegedy, C., Zaremba, W., Sutskever, I., Bruna, J., Erhan, D., Goodfellow, I.,
  and Fergus, R.
\newblock Intriguing properties of neural networks.
\newblock In \emph{ICLR}, 2013.

\bibitem[Tang et~al.(2020)Tang, Huang, and Zhang]{tang2020long}
Tang, K., Huang, J., and Zhang, H.
\newblock Long-tailed classification by keeping the good and removing the bad
  momentum causal effect.
\newblock In \emph{NeurIPS}, 2020.

\bibitem[Tang et~al.(2021)Tang, Tao, and Zhang]{tang2021adversarial}
Tang, K., Tao, M., and Zhang, H.
\newblock Adversarial visual robustness by causal intervention.
\newblock \emph{arXiv preprint arXiv:2106.09534}, 2021.

\bibitem[Teshima et~al.(2020)Teshima, Sato, and Sugiyama]{teshima2020few}
Teshima, T., Sato, I., and Sugiyama, M.
\newblock Few-shot domain adaptation by causal mechanism transfer.
\newblock In \emph{ICML}, 2020.

\bibitem[Vaswani et~al.(2017)Vaswani, Shazeer, Parmar, Uszkoreit, Jones, Gomez,
  Kaiser, and Polosukhin]{vaswani2017attention}
Vaswani, A., Shazeer, N., Parmar, N., Uszkoreit, J., Jones, L., Gomez, A.~N.,
  Kaiser, L., and Polosukhin, I.
\newblock Attention is all you need.
\newblock In \emph{NeurIPS}, 2017.

\bibitem[Weng et~al.(2018)Weng, Zhang, Chen, Song, Hsieh, Daniel, Boning, and
  Dhillon]{weng2018towards}
Weng, L., Zhang, H., Chen, H., Song, Z., Hsieh, C.-J., Daniel, L., Boning, D.,
  and Dhillon, I.
\newblock Towards fast computation of certified robustness for relu networks.
\newblock In \emph{ICML}, 2018.

\bibitem[Wong \& Kolter(2018)Wong and Kolter]{wong2018provable}
Wong, E. and Kolter, Z.
\newblock Provable defenses against adversarial examples via the convex outer
  adversarial polytope.
\newblock In \emph{ICML}, 2018.

\bibitem[Yang et~al.(2019)Yang, Liu, Chen, Ma, and Tsai]{yang2019causal}
Yang, C.-H.~H., Liu, Y.-C., Chen, P.-Y., Ma, X., and Tsai, Y.-C.~J.
\newblock When causal intervention meets adversarial examples and image masking
  for deep neural networks.
\newblock In \emph{ICIP}. IEEE, 2019.

\bibitem[Ye et~al.(2020)Ye, Gong, and Liu]{ye2020safer}
Ye, M., Gong, C., and Liu, Q.
\newblock Safer: A structure-free approach for certified robustness to
  adversarial word substitutions.
\newblock In \emph{ACL}, 2020.

\bibitem[Yue et~al.(2020)Yue, Zhang, Sun, and Hua]{yue2020interventional}
Yue, Z., Zhang, H., Sun, Q., and Hua, X.-S.
\newblock Interventional few-shot learning.
\newblock \emph{arXiv preprint arXiv:2009.13000}, 2020.

\bibitem[Zang et~al.(2020)Zang, Qi, Yang, Liu, Zhang, Liu, and
  Sun]{zang2020sememe}
Zang, Y., Qi, F., Yang, C., Liu, Z., Zhang, M., Liu, Q., and Sun, M.
\newblock Word-level textual adversarial attacking as combinatorial
  optimization.
\newblock In \emph{ACL}, 2020.

\bibitem[Zeng et~al.(2021)Zeng, Zheng, Xu, Li, Yuan, and
  Huang]{zeng2021certified}
Zeng, J., Zheng, X., Xu, J., Li, L., Yuan, L., and Huang, X.
\newblock Certified robustness to text adversarial attacks by randomized
  [mask], 2021.

\bibitem[Zhai et~al.(2020)Zhai, Dan, He, Zhang, Gong, Ravikumar, Hsieh, and
  Wang]{Zhai2020MACERAA}
Zhai, R., Dan, C., He, D., Zhang, H., Gong, B., Ravikumar, P., Hsieh, C.-J.,
  and Wang, L.
\newblock Macer: Attack-free and scalable robust training via maximizing
  certified radius.
\newblock In \emph{ICLR}, 2020.

\bibitem[Zhang et~al.(2020)Zhang, Zhang, and Li]{zhang2020causal}
Zhang, C., Zhang, K., and Li, Y.
\newblock A causal view on robustness of neural networks.
\newblock In \emph{NeurIPS}, 2020.

\bibitem[Zhang et~al.(2018)Zhang, Weng, Chen, Hsieh, and
  Daniel]{zhang2018crown}
Zhang, H., Weng, T.-W., Chen, P.-Y., Hsieh, C.-J., and Daniel, L.
\newblock Efficient neural network robustness certification with general
  activation functions.
\newblock In \emph{Advances in Neural Information Processing Systems
  (NuerIPS)}, dec 2018.

\bibitem[Zhang et~al.(2019)Zhang, Chen, Xiao, Gowal, Stanforth, Li, Boning, and
  Hsieh]{zhang2019stable}
Zhang, H., Chen, H., Xiao, C., Gowal, S., Stanforth, R., Li, B., Boning, D.,
  and Hsieh, C.-J.
\newblock Towards stable and efficient training of verifiably robust neural
  networks, 2019.

\bibitem[Zhang et~al.(2021)Zhang, Gong, Liu, Niu, Tian, Han, Sch{\"o}lkopf, and
  Zhang]{zhang2021adversarial}
Zhang, Y., Gong, M., Liu, T., Niu, G., Tian, X., Han, B., Sch{\"o}lkopf, B.,
  and Zhang, K.
\newblock Adversarial robustness through the lens of causality.
\newblock \emph{arXiv preprint arXiv:2106.06196}, 2021.

\bibitem[Zhao et~al.(2018)Zhao, Dua, and Singh]{zhao2017generating}
Zhao, Z., Dua, D., and Singh, S.
\newblock Generating natural adversarial examples.
\newblock In \emph{ICLR}, 2018.

\end{thebibliography}
\bibliographystyle{icml2022}

\newpage
\appendix
\onecolumn
\section{Proof of Theorem \ref{theorem:causal}}
We have:
    \begin{align}
    p(y|do(x)) &= p^{N\nrightarrow X}(y|x)\\
    &=\int p^{N\nrightarrow X}(y,n|x) \, dn\\
     &=\int p^{N\nrightarrow X}(y|x,n)p^{N\nrightarrow X}(n|x) \, dn\\
    &=\int p^{N\nrightarrow X}(y|x,n)p^{N\nrightarrow X}(n) \, dn\\
    &= \int p(y|x,n)  p(n) \, dn.
    \end{align}


\section{Proof of Theorem \ref{semanticsmooth}}

Define $$B_h=\{h|\arg\max_{y'} \mathop{\mathbb{E}}_{n\sim p(n)}[f_{y'}({h}+n)]=\arg\max_{y'} \mathop{\mathbb{E}}_{n\sim p(n)}[f_{y'}(s(x)+n)] \}$$, and $$B_R=\{h|\|s(x)-h\|_2\leq R\}$$, where $R=\frac{\sigma}{2}(\Phi^{-1}(\mathop{\mathbb{E}}_{n\sim p(n)}[f_y(s(x)+n)])-\Phi^{-1}(\max_{y'\neq y} \mathop{\mathbb{E}}_{n\sim p(n)}[f_{y'}(s(x)+n)])$,
 and $\Phi^{-1}$ is the inverse of standard Gaussian c.d.f. By applying Theorem \ref{softsmooth}, we have classification by
$\arg\max_{y'} \mathop{\mathbb{E}}_{n\sim p(n)}[f_{y'}({h}+n)]$
is robust for all ${h}$, s.t., $\|s(x)-{h}\|_2\leq R$, which means $B_R \subset B_h$.

 According to the condition             \begin{align}
                \|s(x)-s(\hat{x})\|_2\leq \hat{R}, \forall \hat{x} \in \mathbb{B}_{\text{adv}}(x),
            \end{align}
and $\hat{R}\leq R$, we have $$s(\mathbb{B}_{\text{adv}}(x)) \subset B_R$$, which means $$s(\mathbb{B}_{\text{adv}}(x)) \subset B_h$$, i.e. $\arg\max_{y'} \mathop{\mathbb{E}}_{n\sim p(n)}[f_{y'}(s(x)+n)]=\arg\max_{y'} \mathop{\mathbb{E}}_{n\sim p(n)}[f_{y'}(s(\hat{x})+n)],\forall \hat{x} \in \mathbb{B}_{\text{adv}}(x)$. Thus we can get the conclusion that classification by
            $\arg\max_{y'} \mathop{\mathbb{E}}_{n\sim p(n)}[f_{y'}(s(\hat{x})+n)]$
            is robust for all $\hat{x} \in \mathbb{B}_{\text{adv}}(x)$.

\section{Theorem for Randomized Smoothing via a Hard Classifier}

Note that our certification process will employ the hard prediction of $f$,
\emph{i.e.}, $\arg\max_{\overline{y}}
            f_{\overline{y}}(s(x)+n)$,
to make hypothesis test easier to implement.
The certification process is based on the following theorem (which is achieved by  fitting \citet{cohen2019certified} to the semantic smoothing setting).

\begin{theorem}(Robustness by Semantic Smoothing (Hard Classifier Certification))
    \label{hardsemanticsmooth}
        Given $x,y$ and a base classifier $f(s(x)+n)$, where $s(\cdot)$ denotes an encoder that maps input into a semantic space
        and $p(n) \sim N(0,\sigma^2I)$.
        Define hard classifier $g(\cdot)$ as:
        \begin{align}
            g(x) = arg\max_{\overline{y}} f_{\overline{y}}(s(x)+n).
        \end{align}
        If 
        \begin{align}
            \mathop{ \mathbb{P}} 
            \big[
            g(x) = y ]
            \geq \underline{p_A}, \underline{p_A}\in (\frac{1}{2},1],
        \end{align}
        and  $s(\cdot)$ satisfies
            \begin{align}
                \|s(x)-s(\hat{x})\|_2\leq \hat{R}, \forall \hat{x} \in \mathbb{B}_{\text{adv}}(x),
            \end{align}
            then 
            classification by
            $\arg\max_{y'} 
            \mathop{ \mathbb{P}} 
            \big[
            g(x) =y'
            \big]$
            returns $y$ for all $\hat{x}$, if  $\hat{R}\leq R=\sigma(\Phi^{-1}(\underline{p_A}))$
             and $\Phi^{-1}$ is the inverse of standard Gaussian c.d.f.
            \label{latent certi hard}
            \end{theorem}

\section{Proof of Remark 3}
Following the proof technique in \citet{Zhai2020MACERAA},
if  $\gamma\geq \frac{1}{m}$, we have:
\begin{align}
        \mathcal{L}_{\text{cls}} +  \gamma \mathcal{L}_{\text{robust}} & =  \mathop{\mathbb{E}}_{x,y\sim p(x,y)}\mathop{\mathbb{E}}_{n\sim p(n)}-\log  f_y(s(x)+n) +  \gamma *  \mathop{\mathbb{E}}_{x,y\sim p(x,y)} \max(0, \hat{R}-R+m) \\
        &\geq  \mathop{\mathbb{E}}_{x,y\sim p(x,y)} \mathop{\mathbb{E}}_{n\sim p(n)} -\log  f_y(s(x)+n) \\
        &+ \gamma *  \mathop{\mathbb{E}}_{x,y\sim p(x,y)}\max(0, \hat{R}-R+m) * \mathbbm{1}(arg\max_{c\in Y}\mathop{\mathbb{E}}_{n\sim p(n)}f_y(s(x)+n) = y) \\
        &\geq \mathop{\mathbb{E}}_{x,y\sim p(x,y)}\mathbbm{1}(arg\max_{c\in Y}\mathop{\mathbb{E}}_{n\sim p(n)}f_y(s(x)+n) \not = y) \\
        & + \frac{1}{m} *  \mathop{\mathbb{E}}_{x,y\sim p(x,y)}  \max{(0,\hat{R}-R+m)} * \mathbbm{1}(arg\max_{c\in Y}\mathop{\mathbb{E}}_{n\sim p(n)}f_y(s(x)+n) = y)\\
        &\geq \mathop{\mathbb{E}}_{x,y\sim p(x,y)}\mathbbm{1}(arg\max_{c\in Y}\mathop{\mathbb{E}}_{n\sim p(n)}f_y(s(x)+n) \not = y) \\
        & + \mathop{\mathbb{E}}_{x,y\sim p(x,y)}\mathbbm{1}(arg\max_{c\in Y}\mathop{\mathbb{E}}_{n\sim p(n)}f_y(s(x)+n) = y, R-\hat{R}\leq0) \\
        &\geq \mathop{\mathbb{E}}_{x,y\sim p(x,y)}\mathbbm{1}(R-\hat{R}\leq0) \\
        & = 1 - \mathop{\mathbb{E}}_{x,y\sim p(x,y)}\mathbbm{1}(R-\hat{R}>0).
\end{align}

\section{Details of \emph{PvalueBinom} and \emph{LowerConfBound}}\label{statis_detial}

We use the same statistic testing method as the \cite{cohen2019certified}. 

\emph{PvalueBinom}($\text{cnt}_A$, $\text{cnt}_A$ + $\text{cnt}_B$,$p$) calculates the p-value of the two-sided hypothesis test that $\text{cnt}_A$ $\sim$ Binomial($\text{cnt}_A$ + $\text{cnt}_B$, $p$), which is implied as \emph{scipy.stats.binom test}($\text{cnt}_A$, $\text{cnt}_A$ + $\text{cnt}_B$,$p$). 

\emph{LowerConfBound}($\text{cnt}_A, t_2, 1-\alpha $) calculates the one-sided $(1-\alpha)$ lower confidence interval for the Binomial parameter $p$ given that $\text{cnt}_A \sim \text{Binomial}(t_2, p)$, and is implied as \emph{statsmodels.stats.proportion.proportion$\_$confint}($\text{cnt}_A, t_2$, alpha=$2*\alpha$, method="beta")[0].

\section{Trade off between significance criterion $\alpha$ and sampling number $t_2$}\label{alpha300l}

The p-value of a statistical test is strongly correlated with the sample number. Our main results use the sampling number $t_2=30000$ in the second step of certification, which can get p-value lower than the significance criterion of $\alpha=0.001$. However, this consumes around 12 hours to complete the certification using a Tesla V100 for IMDB test set of size 25000. Therefore, in the ablation experiment, in order to improve the efficiency of certification, we use $t_2=300$, and adjust $\alpha$ to 0.05. The comparison table \ref{tab:comp} below  shows that such a setup can achieve similar certification robustness to the main results.

   \begin{table}[htbp]\centering
    
        \begin{tabular}{lcc}
        \hline
        \hline
                               Method  & YELP  & IMDB  \\ \hline

        CISS (30000,0.001)                  & \textbf{90.58} & \textbf{76.45} \\ \hline
        CISS (300,0.05)                  & {90.47} & {75.25} \\
        
        \hline
        \hline
        \end{tabular}
    
    \caption{Comparison between different certification settings.
    }
    \label{tab:comp}
    \end{table}

\end{document}